\documentclass{article} 
\usepackage{iclr2025_conference,times}

\usepackage{amsmath,amsfonts,bm}

\def\eqref#1{equation~\ref{#1}}

\def\1{\bm{1}}

\DeclareMathAlphabet{\mathsfit}{\encodingdefault}{\sfdefault}{m}{sl}
\SetMathAlphabet{\mathsfit}{bold}{\encodingdefault}{\sfdefault}{bx}{n}

\usepackage{hyperref}
\usepackage{url}
\usepackage{url}            
\usepackage{booktabs}       
\usepackage{amsfonts}       
\usepackage{nicefrac}       
\usepackage{microtype}      
\usepackage{xcolor}         
\usepackage{soul}
\usepackage{multirow}
\usepackage{enumerate}

\usepackage{xcolor}         
\usepackage{multirow}
\usepackage{colortbl}
\usepackage{amssymb}
\usepackage{amsmath}
\usepackage{fontawesome}
\usepackage{graphicx}
\usepackage{subcaption}
\usepackage{xr}
\usepackage{array}
\newcolumntype{?}{!{\vrule width 1pt}}
\usepackage{ntheorem}
\usepackage{soul}

\usepackage{minitoc}

\theoremstyle{plain}
\theorembodyfont{}
\theoremsymbol{}
\theoremprework{}
\theorempostwork{}
\theoremseparator{.}

\usepackage[vlined, ruled, linesnumbered]{algorithm2e}

\usepackage{enumitem}

\usepackage{marginnote}

\usepackage[many]{tcolorbox}
\usepackage[frozencache,cachedir=minted-cache]{minted}
\tcbuselibrary{listings,breakable,xparse,skins}

\usepackage[nameinlink,capitalise]{cleveref}
\crefname{section}{Sec.}{Sec.}
\crefname{line}{line}{§§}
\crefname{figure}{Fig.}{Fig.}
\crefname{table}{Tab.}{§§}
\crefname{algorithm}{Alg.}{§§}
\crefname{appendix}{Appx.}{§§}
\crefname{definition}{Def.}{§§}
\crefname{equation}{Eq.}{§§}
\crefname{remark}{Remark.}{§§}

\newcolumntype{\CeX}{>{\centering\let\newline\\\arraybackslash}X}
\newcolumntype{\CeP}{>{\raggedleft\arraybackslash}p}

\usepackage{thmtools} 
\usepackage{thm-restate}

\title{TensorGPT: Efficient Compression\\ of Large Language Models based on\\ Tensor-Train Decomposition}

\iclrfinalcopy
\author{Mingxue Xu, Yao Lei Xu \& Danilo P. Mandic\\ 
Department of Electrical and Electronic Engineering\\
Imperial College London\\
London, SW7 2AZ \\
\texttt{\{m.xu21, yao.xu15, d.mandic\}@imperial.ac.uk}
}

\begin{document}

\maketitle

\begin{abstract}
High-dimensional token embeddings underpin Large Language Models (LLMs), as they can capture subtle semantic information and significantly enhance the modelling of complex language patterns. 
However, this high dimensionality also introduces considerable model parameters and prohibitively high model storage and memory requirements, which is particularly unaffordable for low-end devices.
Targeting no extra training data and insufficient computation cases, we propose a {\bf training-free} model compression approach based on the Tensor-Train Decomposition (TTD), 
whereby each pre-trained token embedding is converted into a lower-dimensional Matrix Product State (MPS).
We then comprehensively investigate the low-rank structures extracted by this approach, in terms of the compression ratio, the language task performance, and latency on a typical low-end device (i.e. Raspberry Pi). 
Taking GPT family models (i.e. GPT-2 and CerebrasGPT) as case studies, our approach theoretically results in $46.89\%$ fewer parameters of the entire model, with a compression ratio $39.38\times$ - $65.64\times$ for the embedding layers. With different hyperparameter choices, the model compressed with our approach can achieve a comparable language task performance to the original model with around $2.0\times$ embedding layer compression. This empirically proves the existence of low-rank structure in GPT family models, and demonstrates that about half of the parameters in the embedding layers are redundant.
\end{abstract}

\section{Introduction}\label{sec:intro}
\vspace{-10px}

\begin{figure}[t]
     \centering
     \begin{subfigure}[b]{0.39\textwidth}
         \centering
         \includegraphics[width=\textwidth]{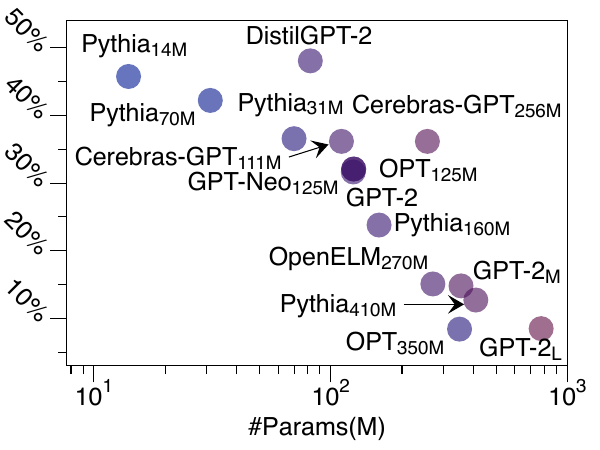}
         \caption{Embedding layer parameter count ratio for common sub-billion LLMs.}
         \label{fig:sub-billion-llms}
     \end{subfigure}
     \hfill
     \begin{subfigure}[b]{0.265\textwidth}
         \centering
         \includegraphics[width=\textwidth]{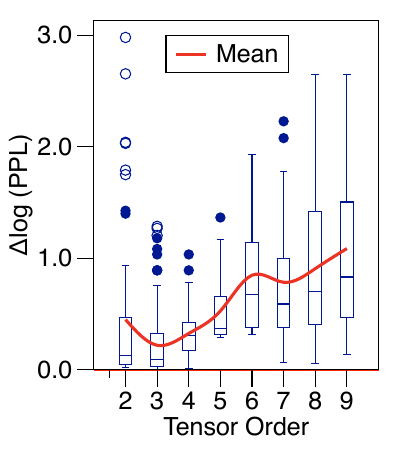}
         \caption{Model performance with different tensor order.}
         \label{fig:box}
     \end{subfigure}
     \hfill
     \begin{subfigure}[b]{0.315\textwidth}
         \centering
         \includegraphics[width=\textwidth]{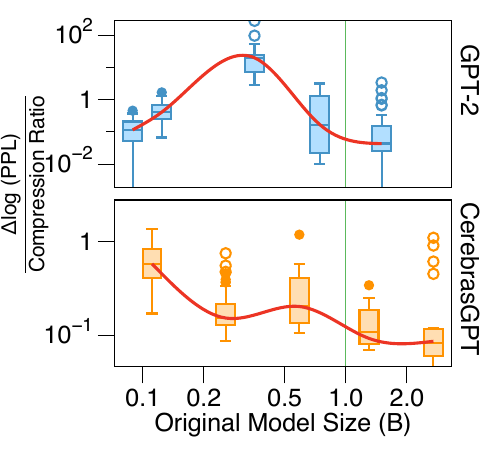}
         \caption{Compression with GPT-2 or CerebrasGPT as original models. }
         \label{fig:points}
     \end{subfigure}
\vspace{-5px}
        \caption{Importance of compressing embedding layers in LLMs and an overview of empirical results of our proposed approach. (a) Embedding layer parameters take a significant proportion of the sub-billion LLMs; in around half of the common sub-billion LLMs, their embedding layer parameters take one-third of the total model parameters.
        (b) Language task performance (we use perplexity, the lower the better) changes of DistilGPT2 when the compression ratio is within $3.0\times$. From our empirical evaluation on 1BW~\citep{chelba2014one}, when the tensor order increases, the language task performance tends to improve first, then decrease after order $3$. 
        (c) The compression trade-off of TensorGPT on different sizes models in GPT-2 and CerebrasGPT series models. This trade-off is roughly measured by the ratio between perplexity and compression ratio of embedding layers; the lower the ratio value, the better the compression. We found that the larger the model size, the better the trade-off, where CerebrasGPT has a smoother trend compared with GPT-2.}
        \label{fig:intro}
        \vspace{-15px}
\end{figure}

Memory and storage efficiency are currently prohibitive to unlocking the full potential of lightweight applications of Large Language Models (LLMs).
Typical LLMs for low-end devices can have less than one billion parameters~\citep{mehta2024openelm,liu2024mobilellm,laskaridis2024melting}, yet they are still a burden for low-end devices.
In the composition of the LLMs parameters, the embedding layers account for a significant portion, especially for sub-billion LLMs~\citep{liu2024mobilellm}. 
As shown in \cref{fig:sub-billion-llms}, for the current typical sub-billion LLMs, the smaller the LLMs, the higher the embedding layer portion, which can be up to $48.08\%$.

Low-rank factorization, which means breaking matrices or higher-dimensional tensors into smaller units, has recently been recognized as a promising solution for LLM efficiency, since its successful application in parameter-efficient fine-tuning~\cite{hu2021lora}. There are already efforts aiming to utilize such a technique in language model compression; some are specialized for embedding layers~\citep{chen2018groupreduce,hrinchuk-etal-2020-tensorized,lioutas2020improving,acharya2019online} while others are not~\citep{chekalina-etal-2023-efficient,chen2021drone,hsu2022language,Dao2022MonarchES,qiu2024compute}. 
However, all of these require an extra training process, such as fine-tuning, meta-learning~\citep{chen2018groupreduce,chen2021drone,hsu2022language,edalati2022kronecker,lioutas2020improving,Dao2022MonarchES,qiu2024compute} and training from scratch~\citep{hrinchuk-etal-2020-tensorized,chekalina-etal-2023-efficient}. There are two limitations to this extra training: 1) extra training involves additional computation and training data, which may be unavailable for low-end devices; 2) training the language model from scratch discards the valuable knowledge stored in the weights of the original models. Given these two limitations, we are wondering: 

{\centering
\ul{\textit{Without extra training, by how much can the embedding layers be compressed without sacrificing language task performance?}}
}

To answer this question, we propose TensorGPT, an approach that induces a high-dimensional tensor structure to capture the information stored in the embedding parameters. 
We use the {\it Tensor-Train Decomposition (TTD)} , which excels in factorizing high-order tensors, to represent embeddings in a lower-rank Matrix Product State (MPS) format. 
Due to the look-up table nature of the embedding layer, we treat each token embedding individually (i.e. each row of the token embedding matrix) rather than taking the token embedding matrix as a whole.
This prevents damaging the individual information of each token, and even applies to the use cases with ever-changing vocabulary. 

We provide a comprehensive discussion of our proposed approach, including general perspectives like compression ratio, language task performance, and more technical perspectives like the impacts of different hyperparameter choices in TensorGPT and the effects of the original model size. 
We found that no matter how we choose the tensor size, which refers to the dimension of each mode of a high-dimensional tensor, the compressed models will finally achieve a comparable language task performance, with a compression ratio $0.5\times$ - $2.0\times$. Some tensor sizes are particularly easier to find reasonable TT ranks, which decide how Tensor-Train Decomposition deal with each tensor mode, to achieve comparable language task performance. For example, in \cref{fig:box}, $3$-order tensors are better at maintaining the task performance compared with the others. Moreover, the larger-scale models have better accuracy-compression trade-offs, as shown in~\cref{fig:points}.

Taking GPT family models as our case study, the contributions of this work are as follows:
 \vspace{-5px}
\begin{enumerate}[leftmargin=*]
    \item As far as we know, we are the first to compress LLMs with low-rank factorization, specifically for low-end devices. We adjust Tensor-Train Decomposition for non-parallel operations of embedding layers, where block-wise approaches~\citep{Dao2022MonarchES,qiu2024compute} are incompetent. 
    \item We test our approach in language modelling and sentiment classification tasks. In both tasks, the compressed models can even outperform the uncompressed models. The larger-scale models generally outperform the uncompressed models in precision and the F1-score in the sentiment classification, and they are more robust in language modelling when the compression ratio increases.  
    \item We provide an ablation study on the general technical aspects of our approach. We measured the latency of various compressed cases of sub-billion GPT models on the low-end Raspberry Pi 5\footnote{\url{https://www.raspberrypi.com/products/raspberry-pi-5/}}, and give a detailed systematic analysis of running TensorGPT on low-end edge devices like Raspberry Pi and higher-end servers.
\end{enumerate}
\section{Preliminaries}

\begin{figure}
    \centering
    \includegraphics[width=\textwidth]{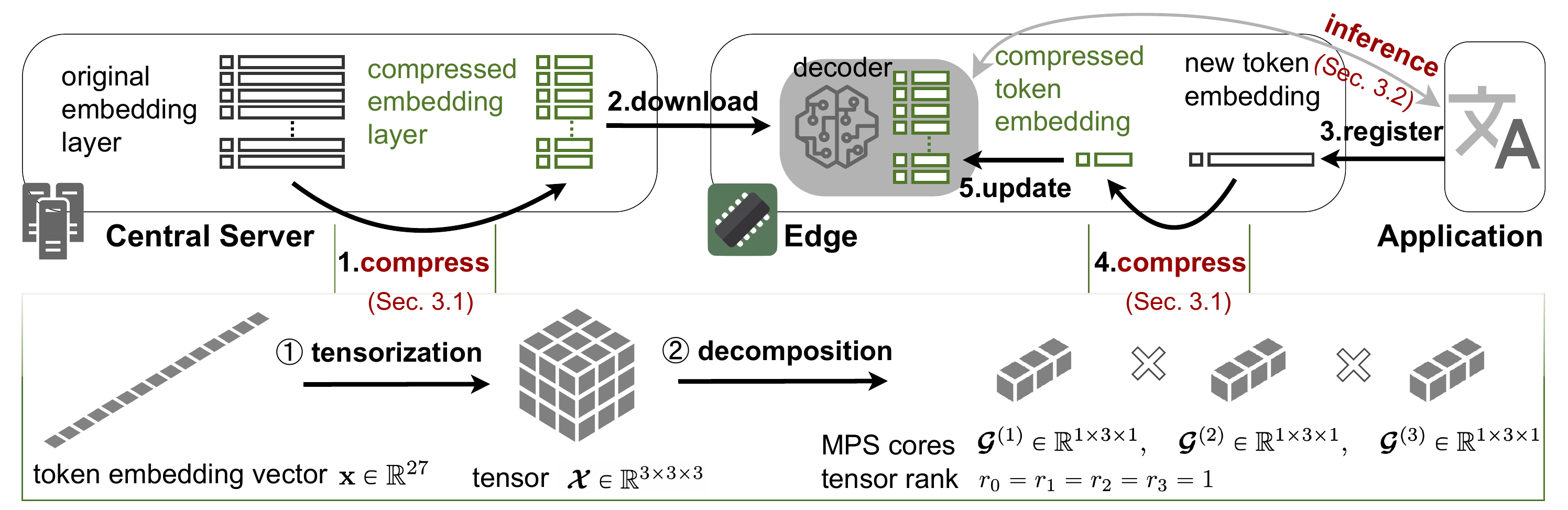}
    \caption{The execution of TensorGPT in edge computing scenario. The central server (server in public or private cloud services, or higher-end personal computer) compresses the token embedding vectors of the whole vocabulary with TensorGPT (step 1). Then, the compressed vocabulary and other parts of the language model (e.g. the decoder) are downloaded and deployed (step 2) on a low-end device (e.g. Raspberry Pi). During the application runs, a new token might be required according to the actual requirements, and registered by the service on the edge device (step 3). The low-end device then compresses this added token embedding with TensorGPT (step 4), and updates (step 5) the current vocabulary of the language model. The compression process of a single token embedding follows a pipeline of tensorization and Tensor-Train Decomposition.}\label{fig:pipeline}
    \vspace{-10px}
    
\end{figure}

This section gives the essential concepts related to tensor, tensor operations and Tensor-Train Decomposition. A more complete introduction about tensors can be found in \cref{sec:preliminaries}.
\paragraph{Order-N Tensor.} An order-$N$ real-valued tensor, $\mathcal{A}$, is a high-dimensional matrix (or multi-way array), denoted by $\mathcal{A}\in\mathbb{R}^{I_1\times\dots\times I_N}$, where $N$ is the order of the tensor (i.e., number of its modes), and $I_k$ ($1 \leq k \leq N$) is the size (i.e., the dimension) of its $k$-{th} mode. In this sense, matrices (denoted as $\mathbf{A}\in\mathbb{R}^{I_1\times I_2}$) can be seen as order-$2$ tensors ($N=2$), vectors (denoted as ${\bf a}\in\mathbb{R}^{I}$) can be seen as order-$1$ tensors ($N=1$), and scalars (denoted as $a\in\mathbb{R}$) are order-$0$ tensors ($N=0$).

\paragraph{Tensorization.} 
A vector ${\bf a} = (a_1, a_2,\ldots,a_{I_1 I_2 \cdots I_N}) \in \mathbb{R}^{I_1 I_2 \cdots I_N}$, can be tensorized (or ``folded'', ``reshaped'') into an order-$N$ tensor $\mathcal{A} \in \mathbb{R}^{I_1 \times I_2 \times \cdots \times I_N}$, so that
\begin{equation}\label{ttcp_eq:folding}
    \mathcal{A}[i_1, i_2, \dots, i_N] = a_{1+\sum_{k=1}^{N}(i_k-1)\prod_{p=1}^{k-1}I_p},\quad\quad 1\leq i_k \leq I_k,
\end{equation}
where $\mathcal{A}[i_1, i_2, \dots, i_N]$ denotes the $(i_1, i_2, \dots, i_N)$-th entry of tensor $\mathcal{A}$.

\paragraph{Vectorization.}\label{para:vec}
Given an order-$N$ tensor, $\mathcal{A}\in\mathbb{R}^{I_1\times\cdots\times I_N}$, its vectorization reshapes the high-dimensional matrix into a vector, $\texttt{vec} \left( \mathcal{A} \right) = \mathbf{\bf{a}} \in \mathbb{R}^{I_1 \cdots I_N}$. 

\paragraph{Tensor Contraction.} 
The contraction of $\mathcal{A}\in\mathbb{R}^{I_1\times \dots \times I_N}$ and $\mathcal{B}\in\mathbb{R}^{J_1\times \dots \times J_M}$, over the $k$th and $p$th modes respectively, where $I_k=J_p$ is denoted as $\mathcal{A}\times^{p}_{k} \mathcal{B}$ and results in a tensor $\mathcal{C} \in \mathbb{R}^{I_1 \times \cdots \times I_{k-1} \times I_{k+1} \times \cdots \times I_N \times J_1 \times \cdots \times J_{p-1} \times J_{p+1}  \times \cdots \times J_M}$, with entries 
\begin{equation}\label{ttcp_eq:contraction}\footnotesize
\begin{aligned}
    &\mathcal{C}[i_1,\dots,i_{k-1}, i_{k+1}, \dots, i_N, j_1, \dots, j_{p-1}, j_{p+1}, \dots, j_M ] \\ = &\sum_{q=1}^{I_k} \mathcal{A}[i_1, \dots, i_{k-1}, q, i_{k+1}, \dots, i_N] \mathcal{B}[j_1, \dots, j_{p-1}, q, j_{p+1}, \dots, j_M]
\end{aligned}
\end{equation}

\paragraph{Tensor-Train Decomposition (TTD).} 
The most common Tensor-Train Decomposition~\citep{oseledets2011tensor} formats a tensor into a Matrix Product State (MPS or TT-MPS) form, which applies the Tensor-Train Singular Value Decomposition (TT-SVD) algorithm (described in \cref{app:ttsvd}) to an order-$N$ tensor, $\mathcal{X} \in \mathbb{R}^{I_1 \times I_2 \times \cdots \times I_N}$. This results in $N$ smaller $2$-nd or $3$-rd order tensors, $\mathcal{G}^{(k)} \in \mathbb{R}^{ r_{k-1} \times  I_k \times r_k }$ for $k=1, \dots, N$, such that 
    \begin{equation}\label{ttcp_eq:tt_def_contract}
    \begin{aligned}
        \mathcal{X} \approx \mathcal{G}^{(1)} \times_2^1 \mathcal{G}^{(2)} \times_3^1 \mathcal{G}^{(3)} \times_3^1 \cdots \times_3^1 \mathcal{G}^{(N)}.
    \end{aligned}
    \end{equation}
Tensor $\mathcal{G}^{(1)}, \ldots, \mathcal{G}^{(N)}$ are referred to as the {tensor cores}, while the set $\{ r_0, r_1, \ldots, r_{N} \}$ represents the  {TT-rank} of the TT decomposition ($r_0=r_N=1$).

\section{Methodology}\label{sec:methodology}

\begin{algorithm}[t]
    \SetKwInOut{Input}{Input}
    \SetKwInOut{Output}{Output}
    \SetKwInOut{Initialize}{Initialize}
\caption{\texttt{TT\_SVD}\citep{oseledets2011tensor} for a Single Token Embedding Compression }
        \label{ttcp_alg:ttsvd}
        
        \Input{1. $d$-dimensional token embedding vector ${\bf x}\in \mathbb{R}^{d}$, approximation accuracy $\epsilon$;\\
             2. Tensor dimension $\{I_1, I_2, \ldots,I_N\}$ and TT ranks $\{r_0,r_1,\ldots,r_N\}$
        }
    
        \Output{
            TT cores $\mathcal{G}^{(1)}, \dots, \mathcal{G}^{(N)}$
        }

        \Initialize{Tensor $\mathcal{X}\leftarrow \texttt{reshape}({\bf x}, [I_1, I_2, \ldots,I_N])$, \\temporary matrix $\textbf{Z} \leftarrow \texttt{reshape}(\mathcal{X},[r_0I_1, \prod_{j=2}^{N} I_j])$, \\truncation parameter $\delta = \frac{\epsilon}{\sqrt{N-1}} \Vert\mathcal{X}\Vert_F$}
    
    
        
        \For{$k=1$ to $N-1$}{

            $\mathbf{U}, \mathbf{S}, \mathbf{V}, \mathbf{E} \leftarrow \texttt{truncSVD}(\textbf{Z},\delta,r_k)$ \label{line:svd}
            \tcp*[f]{\small s.t. $\mathbf{U} \in \mathbb{R}^{r_{k-1}I_k \times r_k}$, ${\Vert \mathbf{E} \Vert_F} \leq \delta$} 
    
            $\mathcal{G}^{(k)} \leftarrow$
                \texttt{reshape} {$\left( \mathbf{U}, [ r_{k-1}, I_k, r_k   ] \right)$}\tcp*[f]{\small get $k$th TT core}\label{line:core}
    
            $\mathbf{Z} \leftarrow$
                \texttt{reshape} {$\left( \mathbf{SV}^T, [r_k I_{k+1}, \prod_{j=k+2}^{N} I_j]) \right)$}\tcp*[f]{\small $\mathbf{SV}^T \in \mathbb{R}^{\prod^{N}_{i=k+2}I_i}$}\label{line:iterate}
        }
    
        $\mathcal{G}^{(N)} \leftarrow \mathbf{Z}$
    
        \Return{
             $\mathcal{G}^{(1)}, \mathcal{G}^{(2)}, \dots, \mathcal{G}^{(N)}$
        } 
    \end{algorithm}
    \vspace{-10px}

This section clarifies the technical cornerstones of our approach. A practical pipeline of our approach is depicted in~\cref{fig:pipeline}. The whole vocabulary is processed on higher-end servers, while inference and vocabulary update happens on lower-end edge devices.

\subsection{Individual Embedding Vector Compression}\label{sec:med-1}

For the compression of the embedding matrix, rather than decomposing the whole embedding weight matrix, we propose to decompose each embedding vector. The lower half of~\cref{fig:pipeline} is a simplified illustration of such a process, with a detailed description in \cref{ttcp_alg:ttsvd}. 

\paragraph{Tensorization.} Each token embedding ${\bf x} \in \mathbb{R}^d$ is reshaped (or folded, tensorized, as in \cref{app:ttsvd}) into an order-$N$ tensor.
Denote $\texttt{reshape}(\cdot)$ as the reshape function, $\mathcal{X}=\texttt{reshape}({\bf x}, \{I_1, I_2, \ldots, I_N\})$ and $\mathcal{X} \in \mathbb{R}^{I_1 \times \cdots \times I_N}$ such that $d = \prod_{k=1}^{N} I_k$. In the example in \cref{fig:pipeline}, the token embedding vector ${\bf x}$ is a $27$-dimensional vector, $d=27$. In this way, vector ${\bf x}$ is reshaped into an order-$3$ ($N=3$) tensor $\mathcal{X}$, with tensor size for each mode $I_1=I_2=I_3=3$. 

\paragraph{Tensor Decomposition.} Tensor $\mathcal{X}$ is then decomposed and stored in a Matrix Product State (MPS) form as $\mathcal{X} \approx \mathcal{G}^{(1)} \times_3^1 \cdots \times_3^1 \mathcal{G}^{(N)}$, with hyperparameters as TT ranks $r_0, r_1, \ldots, r_N$. For the case in \cref{fig:pipeline}, the MPS cores are $\mathcal{G}^{(1)}$, $\mathcal{G}^{(2)}$, $\mathcal{G}^{(3)}$, with TT ranks $r_0=r_1=r_2=r_3=1$.
In other words, instead of storing the entire token embedding vector $\textbf{x} \in \mathbb{R}^{d}$, we store the corresponding MPS cores, $\mathcal{G}^{(k)} \in \mathbb{R}^{r_{k-1} \times I_k \times r_k}$, for $k=1,\ldots,N$. The parameter count of the MPS cores $\{\mathcal{G}^{(k)}\}$ is $\sum^{N}_{k=1} \vert \mathcal{G}^{(k)} \vert = \sum^{N}_{k=1} r_{k-1}I_k r_{k}$, where $\vert \cdot \vert$ represents the parameter count. 

A more detailed explanation of individual token embedding compression is given in \cref{ttcp_alg:ttsvd}, and its cornerstone $\texttt{TT\_SVD}$ is further described in \cref{alg:ttsvd} (in \cref{app:ttsvd}), where $\Vert \cdot \Vert_{F}$ denotes the Frobenius norm. Although the embedding vector is reshaped into a tensor, the decomposition for each mode of this tensor is still based on the matrix-level SVD (\cref{line:svd}). Then the complexity of $\texttt{TT\_SVD}$ can be derived from SVD and its variants, such as truncated SVD~\citep{oseledets2011tensor}. Given the vocabulary size $V$, the original parameters of the embedding layers are compressed from $Vd$ to $V\sum_{k=1}^{N} r_{k-1}I_k r_{k}$, and the compression ratio can be obtained via $\eta_{\texttt{TTD}} =\frac{d}{\sum_{k=1}^{N} r_{k-1}I_k r_{k}} -1$. The computation and memory complexities for all the above processes are summarized in \cref{tab:complexity}.

\begin{table}[]\small
\vspace{-5px}
\caption{Computation and memory complexity during the compression (\cref{sec:med-1}) and inference process (\cref{sec:med-2}) of TensorGPT. $\mathcal{M}_{\texttt{trans}}$ is the transformer module, $V$ denotes the vocabulary size, $d$ is the original token embedding dimension, and $l$ is the token number of the input text. For simplicity, the dimensions for each mode of the tensor and TT rank are represented as $I$ and $r$, respectively, which yields the highest compression ratio when $r=1$ and $I=2$ (the proof is in \cref{app:2-power}).}\label{tab:complexity}
\centering
\begin{tabular}{c|c|c|c|c}
\toprule
 \multirow{2}{*}{{\bf Memory}}& Original & Compressed & Encoded Texts & Input to $\mathcal{M}_{\texttt{trans}}$ \\ \cline{2-5}
 &  $\mathcal{O}(Vd)$      & $\mathcal{O}(VNIr^3)$  &$\mathcal{O}(lNIr^2)$  &      $\mathcal{O}(ld)$                          \\\midrule
  \multirow{2}{*}{{\bf Computation}}& \multicolumn{2}{c|}{$\texttt{TT-SVD}$}  &  \multicolumn{2}{c}{Reconstruction}   \\\cline{2-5}
  & \multicolumn{2}{c|}{$\mathcal{O}(NIr^3)$}         & \multicolumn{2}{c}{$\mathcal{O}(NIr^2)$}            \\   
  \bottomrule
\end{tabular}
\vspace{10px}

\end{table}

\subsection{Language Model Inference Process with the Compressed Embeddings}\label{sec:med-2}

The original inference process with embedding vectors is as follows: when the encoded texts (separated as tokens) are forwarded to the embedding layer, the embedding layer outputs the embedding vectors according to the input tokens; the embedding layer here acts like a look-up table. The embedding vectors are then forwarded to the hidden layers of the transformer, whose size is the same as the dimension of the embedding vectors. Thus, if there is no internal change in the hidden layers, the dimension of the embedding vectors should compile with the dimension of the hidden layers. So, the compressed embeddings should be reconstructed to the original dimension to enable the forwarding process. This inference happens at the application phase shown in the upper right part of~\cref{fig:pipeline}.

Thus just before forwarding embedding vectors to the hidden layers, the memory usage increases from $l\sum_{k=1}^{N} r_{k-1}I_k r_{k}$ to $ld$. However, given that the vocabulary size $V$ is normally much larger than the input token number $l$, that means $V\gg l$. Thus our approach can still significantly reduce the memory usage if the embedding layer takes a significant part of the whole model parameters. The reconstruction process follows the tensor contraction in \cref{ttcp_eq:contraction}, turning the TT cores $\{\mathcal{G}^{(k)}\}$ into a $N$-order tensor $\mathcal{X}$ according to \cref{ttcp_eq:tt_def_contract}, and then vectorizing $\mathcal{X}$ into a full-size embedding vector according to \cref{para:vec}.

\section{Empirical Evaluation}\label{sec:exp}

\subsection{Experimental Setup} 
\paragraph{Models, Tasks and Dataset.} The sub-billion models we used in the GPT family are DistilGPT2~\citep{sanh2019distilbert}, GPT2, GPT2-M, GPT2-L~\citep{radford2019language}, CerebrasGPT-111M, CerebrasGPT-256M and CerebrasGPT-590M~\citep{dey2023cerebras}. We also tested the models of over a billion parameters for language task performance with GPT2-XL (1.5 billion parameters) and CerebrasGPT-1.3B (1.3 billion parameters). The evaluated language tasks were language modelling and sentiment classification. For language modelling, the considered datasets are WikiText2, WikiText103~\citep{merity2022pointer} and 1BW~\citep{chelba2014one}. For sentiment classification, the considered dataset is IMDB~\citep{maas-EtAl:2011:ACL-HLT2011}. 

\paragraph{Hardware.} Our main experiments were completed on a GPU workstation with an RTX A6000 48GB GPU and AMD Ryzen Threadripper PRO 5955WX CPU. The GPU resource was mainly used to fine-tune language modelling models for sequence classification, which is the requirement of the sentiment classification task. The inference latency of the low-end devices was measured on a Raspberry Pi 5, with a 64-bit Arm Cortex-A76 CPU and 8GB SDRAM. 

\subsection{Evaluation Metrics}\label{sec:metrics}
\paragraph{Compression Ratio.} Denote $\mathcal{M}$ as a model block set containing a list of model modules like embedding layers and attention layers. With $\mathcal{M}_0$ as the original model block set, $\mathcal{M}_\texttt{cmpr}$ as the compressed version of $\mathcal{M}_0$, and $\vert\mathcal{M}\vert$ as the parameter count of $\mathcal{M}$. The compression ratio $\eta$ is defined as  
\begin{equation}\label{def:cr}
  \eta = \frac{\vert \mathcal{M}_0 \vert - \vert\mathcal{M}_\texttt{cmpr}\vert}{\vert \mathcal{M}_\texttt{cmpr} \vert}. 
\end{equation}
Specifically, the embedding compression rate is $\eta_{\texttt{emb}} = \frac{\vert \mathcal{T}_0 \vert - \vert\mathcal{T}_\texttt{cmpr}\vert}{\vert \mathcal{T}_0 \vert}$, where $\mathcal{T}$ only contains token embedding layer and position embedding layer. 

\paragraph{Perplexity and Logarithmic Perplexity.} Perplexity is used as a performance evaluation metric of the language modelling task, which has the following form
\begin{equation}\label{eq:ppl}\small
    \texttt{PPL}(S,\mathcal{M}) = \left( \prod^{|S|}_{i=1} p_{\mathcal{M}}(x_i | x_1, x_2, \ldots, x_{i-1}) \right)^{-1}
\end{equation}
where $S$ is an ordered set (token sequence), consisting of a set of tokens $\{x_t\}$, $t=1,2,\ldots, |S|$, and $\mathcal{M}$ is the model block that contains all the modules of the language model we evaluate. 

Notice that the compression ratio \cref{def:cr} has a linear form, while perplexity \cref{eq:ppl} has an exponential form, so it is hard to combine them as a description of a model compression result, since when compression ratio $\eta$ linearly increases, the perplexity $\texttt{PPL}$ explodes exponentially. To this end, we use the following logarithmic form to describe the language modeling performance
\begin{equation}\label{eq:ppl-ln}\small
    \ln{\texttt{PPL}(S,\mathcal{M})} = - \sum ^{|S|}_{i=1} \ln p_{\mathcal{M}}(x_i | x_1, x_2, \ldots, x_{i-1})
\end{equation}

Now, the language modelling performance change before and after compression is given by  
\begin{equation}\small\label{def:logppl}
\Delta \ln{\texttt{PPL}(S,\mathcal{M})} = \ln{\texttt{PPL}(S,\mathcal{M_{\texttt{cmpr}}})} - \ln{\texttt{PPL}(S,\mathcal{M}_0)} = \sum ^{|S|}_{i=1} \ln \frac{p_{\mathcal{M}_{0}}(x_i | x_1, x_2, \ldots, x_{i-1})}{p_{\mathcal{M}_{\texttt{cmpr}}}(x_i | x_1, x_2, \ldots, x_{i-1})},
\end{equation}
observe that \cref{def:logppl} exhibits linearity, like \cref{def:cr}.

\paragraph{Accuracy, Precision, Recall and F1-Score.} We use these four common evaluation metrics for classification to comprehensively analyze the classification performance of the compressed model. To investigate the performance change before and after compression, we use the difference between the metric values after and before the compression.

\begin{figure}[t]
     \centering
     \begin{subfigure}[b]{0.245\textwidth}
         \centering
         \includegraphics[width=\textwidth]{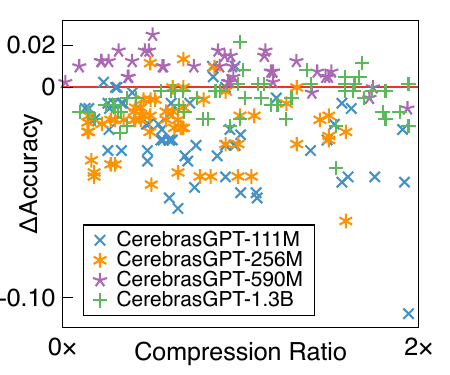}
         \caption{Accuracy}
         \label{fig:cls-acc}
     \end{subfigure}
          \hfill
     \begin{subfigure}[b]{0.245\textwidth}
         \centering
         \includegraphics[width=\textwidth]{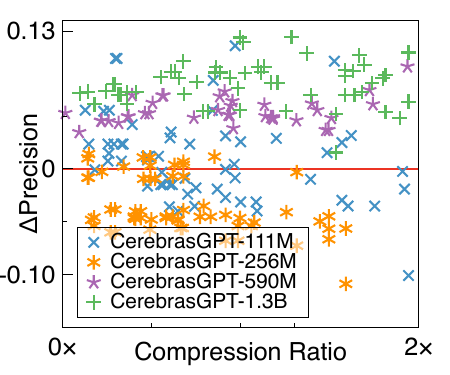}
         \caption{Precision}
         \label{fig:cls-prec}
     \end{subfigure}
     \hfill
          \begin{subfigure}[b]{0.245\textwidth}
         \centering
         \includegraphics[width=\textwidth]{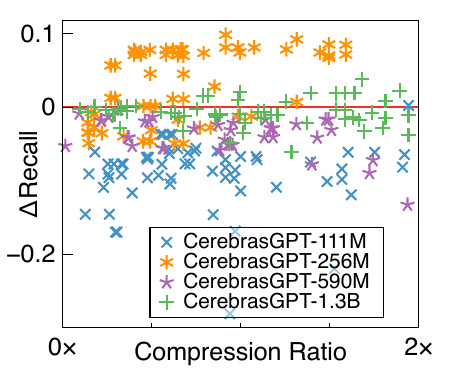}
         \caption{Recall}
         \label{fig:cls-recall}
     \end{subfigure}
     \hfill
    \begin{subfigure}[b]{0.245\textwidth}
         \centering
         \includegraphics[width=\textwidth]{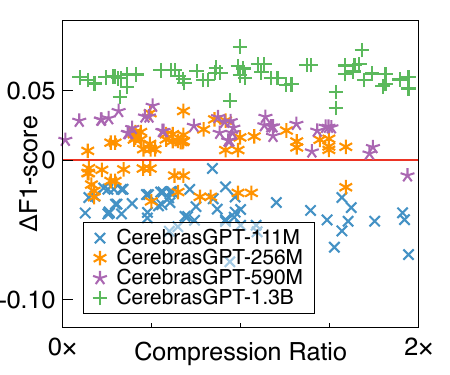}
         \caption{F1-Score}
         \label{fig:cls-f1}
     \end{subfigure}
     \hfill
     \begin{subfigure}[b]{0.245\textwidth}
         \centering
         \includegraphics[width=\textwidth]{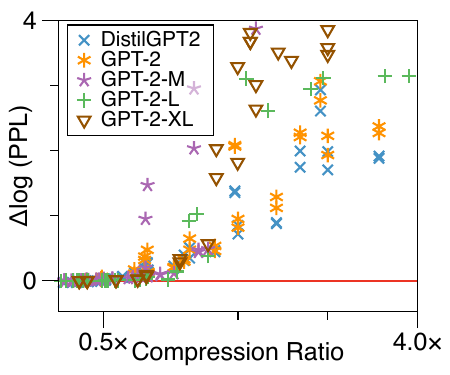}
         \caption{Tensor order $N=2$}
         \label{fig:lnppl-1}
     \end{subfigure}
          \hfill
     \begin{subfigure}[b]{0.245\textwidth}
         \centering
         \includegraphics[width=\textwidth]{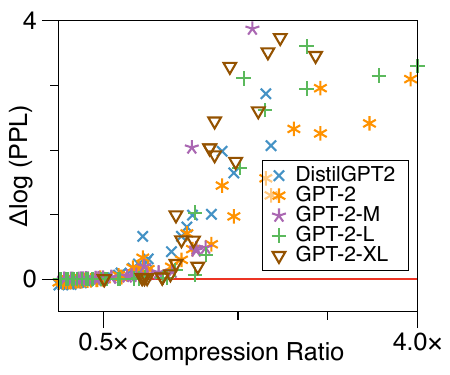}
         \caption{Tensor order $N=3$}
         \label{fig:lnppl-2}
     \end{subfigure}
     \hfill
          \begin{subfigure}[b]{0.245\textwidth}
         \centering
         \includegraphics[width=\textwidth]{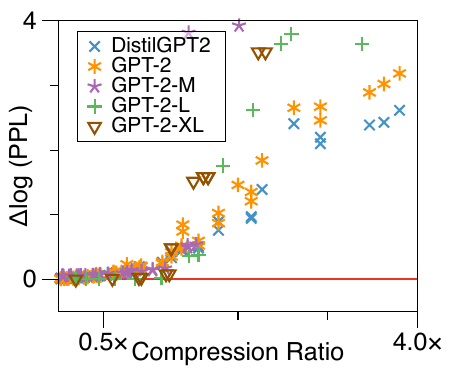}
         \caption{Tensor order $N=4$}
         \label{fig:lnppl-3}
     \end{subfigure}
     \hfill
    \begin{subfigure}[b]{0.245\textwidth}
         \centering
         \includegraphics[width=\textwidth]{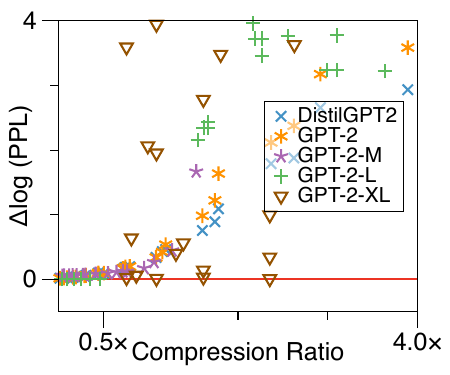}
         \caption{Tensor order $N=5$}
         \label{fig:lnppl-4}
     \end{subfigure}
     \hfill
          \centering
     \begin{subfigure}[b]{0.245\textwidth}
         \centering
         \includegraphics[width=\textwidth]{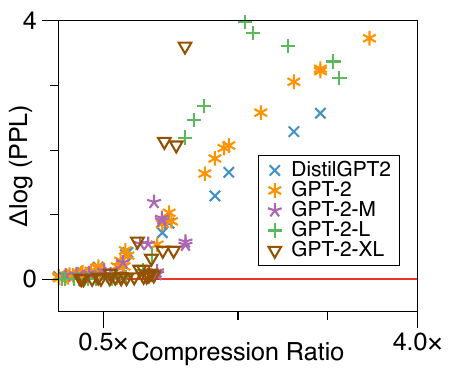}
         \caption{Tensor order $N=6$}
         \label{fig:lnppl-5}
     \end{subfigure}
          \begin{subfigure}[b]{0.245\textwidth}
         \centering
         \includegraphics[width=\textwidth]{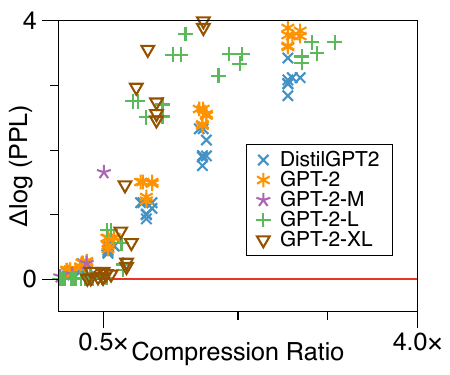}
         \caption{Tensor order $N=N_{\text{max}}$}
         \label{fig:lnppl-6}
     \end{subfigure}
          \begin{subfigure}[b]{0.49\textwidth}
         \centering
         \includegraphics[width=\textwidth]{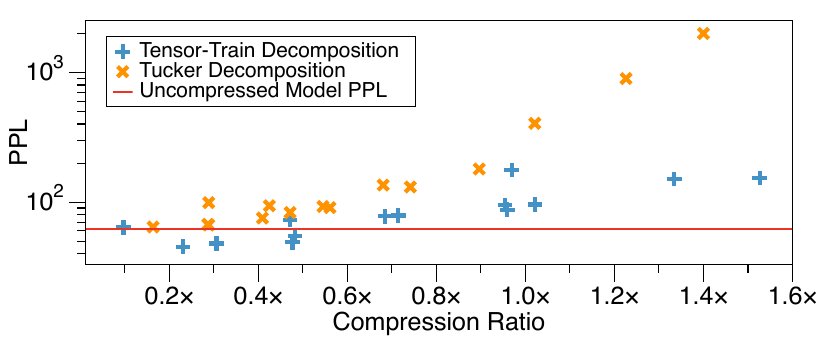}
         \caption{Comparison of different low-rank approaches.}
         \label{fig:tucker}
     \end{subfigure}

        \caption{Language task performance change after the compression. (a)-(d): The accuracy, precision, F1-score and recall of sentiment classification, with the increasing compression ratio of the CerebrasGPT models. The higher the values, the better the classification performance. (e)-(j): The language modelling performance and compression ratio of the GPT-2 series models, where $N$ denotes the tensor order. The best compression cases (higher compression ratio with negligible drop in accuracy) occur when $N=3$, implying the optimal feature representation of the token embedding vectors may be three-dimensional. (g): Different tensor decomposition approaches, where $\text{PPL}$ means perplexity.}
        \label{fig:lnppl}
        \vspace{-10px}
\end{figure}

\subsection{Experimental Results}\label{sec:exp-res}
\paragraph{Compression Ratio and Language Task Performance.} 
The language modelling performance and the compression ratio of different tensor orders and models are shown in \cref{fig:lnppl}. There is no general conclusion as to whether higher-order tensors are better than lower-order tensors, but roughly speaking, compression ratio and language modelling performance in \cref{fig:lnppl-1}-\cref{fig:lnppl-4} are better than those in \cref{fig:lnppl-5}-\cref{fig:lnppl-6}. This implies that there may exist a high-dimensional representation in the weight of embedding layers, no more than six-dimensional and most probably around three-dimensional. On the other hand, due to the combination of tensor size and TT ranks exponentially exploding, we could not test all the possible combinations. However, we can still observe that independent of the tensor orders and the models used for the compression, significant language modelling performance loss tends to appear when the compression ratio exceeds $2.0\times$. We further compared our proposed approach with the Tucker decomposition in~\cref{fig:tucker} with the same tensorization strategy in~\cref{sec:med-1}, and found our adopted Tensor-Train Decomposition outperforms the Tucker Decomposition in perplexity. 

The results of the sentiment classification task also show that the robustness of larger-scale models (Cerebras-590M and Cerebras-1.3B) is better than that of the smaller models (Cerebras-111M and Cerebras-256M), similar to the trend in language modelling tasks mentioned above. The compressed larger-scale models tend to outperform the original model in precision and F1-score, indicating that our compression improves the ability of the larger models to recognize the positive texts. In contrast, the smaller models tends to have worse performance when the compression ratio increases.

A notable observation is that in both language modelling (\cref{fig:lnppl-1,fig:lnppl-2,fig:lnppl-3,fig:lnppl-4,fig:lnppl-5,fig:lnppl-6}) and sentiment classification (\cref{fig:cls-acc,fig:cls-prec,fig:cls-recall,fig:cls-f1}), the larger models (GPT-2-M, GPT-2-L, GPT-2-XL, CerebrasGPT-590M and CerebrasGPT-1.3B) are more robust to the compression ratio increase, compared with smaller models (DistilGPT2, GPT-2, CerebrasGPT-111M and CerebrasGPT-256M), especially when the compression ratio is less than $1.0\times$. 
This is probably because the embedding layers take a smaller proportion of the entire model, and also sheds light that TensorGPT might be used to improve the language task performance for large-scale models.

\paragraph{Latency.} While TensorGPT significantly reduces the model parameters and even improves the language tasks performance, in practice it also introduced latencies - compression latency in~\cref{sec:med-1}  and inference latency~\cref{sec:med-2}. 

For the compression latency, we investigated the compression latency on the token level, as shown in~\cref{tab:decompose-latency}. Here, ``original'' means the uncompressed model, while $\texttt{PPL}_{\alpha}$ means the compressed model with a negligible task performance drop. In our case, ``negligible task performance drop'' means in the language modelling task, the perplexity is no more than $100.0$. The notation $\varphi_{\text{max}}$ refers to the compressed model with maximum compression ratio.
We observed that for individual token embeddings, there was no significant latency difference between high-end servers and Raspberry Pi, typically no more than 2 milliseconds for each token. Thus, it is acceptable for the Raspberry Pi to compress the individual token embeddings. 

For the inference latency of a single text, we chose a typical text length of 50 tokens, as shown in~\cref{tab:exp-profile}. we used ``original'', $\texttt{PPL}_{\alpha}$, $\varphi_{\text{max}}$ same as those in~\cref{tab:decompose-latency}, to represent the uncompressed model, the compressed model with a negligible task performance drop and the model with a maximum compression ratio. A typically induced latency for an input text was no more than $0.3$ seconds, which is acceptable for edge applications. 

\begin{table}[]\scriptsize	
\centering
\caption{The latency (ms/token) of tensorization \& decomposition token embedding vectors and reconstruction on the high-end and lower-end devices. $\texttt{PPL}_{\alpha}$ means the compressed model with a negligible task performance drop, and the symbol $\varphi_{\text{max}}$ represents the case with a maximum compression ratio. $d_{\text{emb}}$ is the embedding dimension of the token embedding vector, and the tested models are GPT-2 and GPT-2-M. On the CPU level, for single token embedding vector decomposition and reconstruction, both server and edge devices have no significant computation overhead.}\label{tab:decompose-latency}
\begin{tabular}{l|l|cc|ll}
\toprule
\multirow{2}{*}{\begin{tabular}{c}
\textbf{Device (CPU)} \\
(ms/token)
\end{tabular}}              & \multirow{2}{*}{$d_{\text{emb}}$} & \multicolumn{2}{c|}{{\begin{tabular}[c]{@{}l@{}}\textbf{tensorization}\\ \& \textbf{decomposition}\end{tabular}}}             & \multicolumn{2}{c}{\textbf{reconstruction}}          \\ \cline{3-6} 
     &                   & \multicolumn{1}{c|}{$\texttt{PPL}_{\alpha}$} & \multicolumn{1}{c|}{$\varphi_{\text{max}}$} & \multicolumn{1}{c|}{$\texttt{PPL}_{\alpha}$} & \multicolumn{1}{c}{$\varphi_{\text{max}}$}  \\ \midrule
\multirow{2}{*}{\textbf{Server}}         &        768           & \multicolumn{1}{l|}{0.627}    &    1.429                          & \multicolumn{1}{l|}{0.117}    &     0.238                   \\ \cline{2-6} 
     &        1024           & \multicolumn{1}{l|}{0.452}    &      1.512                        & \multicolumn{1}{l|}{0.114}    &             0.261             \\ \midrule
\multirow{2}{*}{\textbf{Raspberry Pi 5}} &        768           & \multicolumn{1}{l|}{0.760}    &    1.948                            & \multicolumn{1}{l|}{0.330}    &            0.468       \\ \cline{2-6} 
     &        1024           & \multicolumn{1}{l|}{0.612}    &    2.148                          & \multicolumn{1}{l|}{0.364}    &   0.614                        \\ \bottomrule
\end{tabular}
\end{table}

\begin{table}[]\scriptsize	
    \caption{Parameters, number of floating-point operations (flops) of the compressed and uncompressed sub-billion models, and latency on Raspberry Pi CPU. For flops, the token number of the input texts is $100$, while for latency on Raspberry Pi, the token number is $50$.}\label{tab:exp-profile}
    \centering

\begin{tabular}{ll?llll?lll}
\toprule
\multicolumn{2}{c?}{\multirow{2}{*}{\textbf{GPT Models}}}                                & \multicolumn{4}{c?}{\textbf{GPT2}}                                                                          & \multicolumn{3}{c}{\textbf{CerebrasGPT}}                              \\ \cline{3-9} 
\multicolumn{2}{l?}{}                                                    & \multicolumn{1}{c|}{{\bf DistilGPT2}} & \multicolumn{1}{c|}{{\bf GPT-2}} & \multicolumn{1}{c|}{{\bf GPT-2-M}} & \multicolumn{1}{c?}{{\bf GPT-2-L}} & \multicolumn{1}{l|}{{\bf 111M}} & \multicolumn{1}{l|}{{\bf 256M}} & {\bf 590M} \\ \midrule
\multicolumn{1}{l?}{\multirow{3}{*}{\begin{tabular}[c]{@{}l@{}}\textbf{\# Params}\\ (M)\end{tabular}}} & original               & \multicolumn{1}{l|}{$81.9$}           & \multicolumn{1}{l|}{$124.44$}     & \multicolumn{1}{l|}{$354.82$}       &  $774.03$      &  \multicolumn{1}{l|}{$111.05$}     & \multicolumn{1}{l|}{$255.98$}     &   $590.31$   \\ \cline{2-9} 
\multicolumn{1}{l?}{}                           & $\texttt{PPL}_{\alpha}$                   & \multicolumn{1}{l|}{$67.06$}           & \multicolumn{1}{l|}{$106.36$}     & \multicolumn{1}{l|}{$326.45$}       &   $734.28$     & \multicolumn{1}{l|}{$101.78$}     & \multicolumn{1}{l|}{$226.69$}     &   $543.45$   \\ \cline{2-9} 
\multicolumn{1}{l?}{}                           & $\varphi_{\text{max}}$  & \multicolumn{1}{l|}{$43.45$}           & \multicolumn{1}{l|}{$85.99$}     & \multicolumn{1}{l|}{$303.88$}       &    $710.83$    & \multicolumn{1}{l|}{$71.87$}     & \multicolumn{1}{l|}{$200.59$}     &    $511.07$  \\ \midrule
\multicolumn{1}{l?}{\multirow{3}{*}{\begin{tabular}[c]{@{}l@{}}\textbf{flops} \\ ($10^6$/text )\end{tabular}}}     & original               & \multicolumn{1}{l|}{$20250$}           & \multicolumn{1}{l|}{$40490$}     & \multicolumn{1}{l|}{$142250$}       &    $330980$    & \multicolumn{1}{l|}{$14470$}     & \multicolumn{1}{l|}{$40400$}     &    $103060$  \\ \cline{2-9} 
\multicolumn{1}{l?}{}                           & $\texttt{PPL}_{\alpha}$                    & \multicolumn{1}{l|}{$+ 1.65$}           & \multicolumn{1}{l|}{$+ 1.88$}     & \multicolumn{1}{l|}{$+ 3.11$}       &    $+ 2.30$    & \multicolumn{1}{l|}{$+ 0.38$}     & \multicolumn{1}{l|}{$+ 1.63$}     &   $+ 2.30$   \\ \cline{2-9} 
\multicolumn{1}{l?}{}                           & $\varphi_{\text{max}}$  & \multicolumn{1}{l|}{$+ 0.13$}           & \multicolumn{1}{l|}{$+ 0.13$}     & \multicolumn{1}{l|}{$+ 0.20$}       &    $+ 0.25$    & \multicolumn{1}{l|}{$+ 0.13$}     & \multicolumn{1}{l|}{$+ 0.12$}     &   $+ 0.26$   \\ \midrule
\multicolumn{1}{l?}{\multirow{3}{*}{\begin{tabular}[c]{@{}l@{}}\textbf{Latency on} \\ \textbf{Raspberry} \\ \textbf{Pi} (s/text)\end{tabular}}}   & original               & \multicolumn{1}{l|}{$0.19_{\pm 0.02}$}           & \multicolumn{1}{l|}{$0.50_{\pm 0.19}$}     & \multicolumn{1}{l|}{$1.23_{\pm 0.12}$}       &    $3.01_{\pm 0.47}$    & \multicolumn{1}{l|}{$0.47_{\pm 0.21}$}     & \multicolumn{1}{l|}{$0.71_{\pm 0.02}$}     &   $1.81_{\pm 0.25}$   \\ \cline{2-9} 
\multicolumn{1}{l?}{}                           & $\texttt{PPL}_{\alpha}$                    & \multicolumn{1}{l|}{$0.36_{\pm 0.19}$}           & \multicolumn{1}{l|}{$0.50_{\pm 0.16}$}     & \multicolumn{1}{l|}{$1.26_{\pm 0.22}$}       &   $3.01_{\pm 0.29}$     & \multicolumn{1}{l|}{$0.48_{\pm 0.23}$}     & \multicolumn{1}{l|}{$1.01_{\pm 0.29}$}     &    $1.89_{\pm 0.28}$  \\ \cline{2-9} 
\multicolumn{1}{l?}{}                           & $\varphi_{\text{max}}$  & \multicolumn{1}{l|}{$0.19_{\pm 0.03}$}           & \multicolumn{1}{l|}{$0.71_{\pm 0.38}$}     & \multicolumn{1}{l|}{$1.55_{\pm 0.36}$}       &    $3.52_{\pm 0.44}$    & \multicolumn{1}{l|}{$0.72_{\pm 0.42}$}     & \multicolumn{1}{l|}{$0.95_{\pm 0.27}$}     &   $1.91_{\pm 0.24}$   \\ \bottomrule
\end{tabular}
\end{table}

\section{Related Work}\label{sec:related}

\subsection{Matrix or Tensor Factorization for Language Model Compression}
Low-rank factorization can break the high-dimensional weight matrices into smaller matrices or tensors, so that the overall size of the model can be shrunk. According to the dimensions of the structure that the original weight matrices are broken into, these approaches can be divided into matrix-based and tensor-based.

\paragraph{Matrix-based Approaches.}
A straightforward way to shrink the model size is to decompose weight matrices via singular value decomposition (SVD)~\citep{acharya2019online}, which can be further improved by the weighted approach considering the model performance afterwards~\citep{hsu2022language}, knowledge distillation~\citep{lioutas2020improving,mao2020ladabert} and pruning~\citep{mao2020ladabert}. There are also some block-wise decomposition approaches used in language model compression, like Kronecker Products~\citep{tahaei-etal-2022-kroneckerbert,edalati2022kronecker} and data-driven block-wise partitioning~\citep{chen2018groupreduce,chen2021drone}. 

\citep{Dao2022MonarchES,qiu2024compute} used the block-diagonal matrices to reduce the FLOPs in the linear layers computation, with the bonus of shrinking the model size. However, our paper focuses on reducing the parameters of embedding layers, and there is no monotonous relationship between the FLOPs (computation cost) and parameters (memory usage)~\citep{mcunet}. Also, their investigated matrix multiplication only occurs in feed-forward layers, thus their approaches do not fit the embedding layer compression. Moreover, block-diagonal matrices are optimised for GPUs for better parallelization. Our aim of minimizing the number of parameters, makes it optimized for lower-end edge devices rather than GPUs. Indeed, on Raspberry Pi 5, the additional forwarding latency due to compressed embeddings (0.330 - 0.364ms /token in~\cref{tab:decompose-latency}) is even faster than that on GPU (measured as 0.463ms /token in our setting), since there is no parallelization during this forwarding process.

\paragraph{Tensor-based Approaches.} Despite some efforts to use tensor decomposition to compress the language model size, all come with an extra training process. The works in \citep{abronin2024tqcompressor} use Kronecker decomposition with row-column permutation during the GPT model fine-tuning process, while \citep{hrinchuk-etal-2020-tensorized} and \citep{chekalina-etal-2023-efficient} propose a tensor-train structured embedding layer and GPT model respectively, yet both train the new-structured model from scratch.

\subsection{Tensor Network and Tensor Network Structure Search}
The works in \citep{li2023alternating,li2022permutation} propose a local search technique to solve a model-agnostic objective function to optimize model complexity, with~\citep{li2023alternating} or without~\citep{li2022permutation} a simultaneous tensor shape optimization, and with image tasks (e.g. image compression and completion) as case studies. However, since they do not consider more complex functions like language modelling, which is discussed in our work, this higher complexity might drive the optimization process unsolvable. 
Another branch of tensor network search involves a learning-based approach~\citep{yin2022batude}, which trains a meta-model to learn the patterns of the low-rank pattern of the model parameters.

\section{Conclusion and Future Work} \label{sec:conclusion}
In the context of Large Language Models (LLMs) with a parameter count of less than one billion, the embedding layers take significant proportions of the total model parameters. The low-rank approach is a promising technique for parameter reduction, however, the existing low-rank approaches to compressing LLMs all involve an extra training process. Based on the Tensor-Train Decomposition, this work aims to find the low-rank structure {\it without the need for extra training}, and investigate the compression ratio, language task performance (i.e. language modelling and sentiment classification), and latency on typical low-end edge device Raspberry Pi. We have found that for the larger models (GPT-2-M, GPT-2-L, GPT-2-XL, CerebrasGPT-590M and CerebrasGPT-1.3B), the language task performance was maintained and even improved while reducing the model size. Also, we have obtained an empirical compression bound on our proposed approach, with a $0.5\times$ to $2.0\times$ compression ratio of the embedding layers, without language task performance loss. 
We measured the latency of our approach on a typical low-end device (i.e. Raspberry Pi), with a significant parameter reduction, the on-device compression latency was typically no more than $2$ ms for each token, highly competent for edge applications. 

There are both limitations to our work, and more importantly, rather a broad range of future work following our proposed TensorGPT. Firstly, the tensorized embedding layers do not natively compile with the hidden layers, so tensorized hidden layers are required. Also, although tensor operations, like contraction, do not require much memory, they might need more arithmetic operations on the CPU, thus requiring accelerated tensor operations. Last but not least, this work focuses on GPT-series models and reaches the ``half parameters redundancy'' conclusion. Extensions to other kinds of models, like the LLama family models, are the subject of future work.

\newpage
\bibliography{ref}
\bibliographystyle{iclr2025_conference}
\newpage
\appendix

\begin{center}
    \Large{\bf{Appendix}}
\end{center}
\addcontentsline{toc}{section}{Appendix} 
\parttoc 

\section{Preliminaries}\label{sec:preliminaries}
Our notation in this paper is summarized as follows:

\begin{table}[h]
    \centering
    \begin{tabular}{r|l}
    \toprule
    {\bf Symbol} & {\bf Meaning} \\
    \midrule
        $a$ & Scalar\\
         ${\bf x}$& Vector \\
         ${\bf A}$& Matrix \\
         $\mathcal{X}$, $\mathcal{A}$, $\mathcal{B}$& Tensor \\
         $N$ & Tensor order\\
         $\mathcal{X}[i_1, i_2, \ldots, i_N]$ & The $(i_1, i_2,\ldots,i_N)$th entry of the tensor\\
         $I, I_k$ & Tensor dimension, tensor dimension for the $k$th mode   \\
         $\mathcal{M}$ & Model module set \\
         $\vert \mathcal{M}\vert$, $\vert \mathcal{G}\vert$, $\vert S\vert$ & Parameter count of the model module set $\mathcal{M}$, tensor $\mathcal{G}$ or cardinality of set $S$ \\
         $V$ & Vocabulary of the language model\\
         $d$ & Token embedding dimension \\
         $l$ & Input text length\\
         $r$, $r_k$& TT rank, TT rank of the $k$th mode of the tensor\\
         $\mathcal{G}^{(k)}$& TT(MPS) core of the $k$th mode of the tensor\\
         $\times^{p}_{k}$ & Tensor contraction for the $p$th (formal tensor) and $k$th (latter tensor) mode \\
         $\eta$ & Compression ratio of the entire model\\
         $\eta_{\texttt{emb}}$ & Compression ratio of the embedding layer \\
         $\varphi$ & Parameter reduction ratio of the whole model. \\
         $\varphi_{\texttt{emb}}$ & Parameter reduction ratio of the embedding layer. \\
         
         \bottomrule
    \end{tabular}
    \caption{Notation in this paper.}
    \label{tab:notation}
\end{table}

\begin{algorithm}
        \SetKwInOut{Input}{Input}
        \SetKwInOut{Output}{Output}
        \caption{Tensor-Train Singular Value Decomposition (TT-SVD) }
        \label{alg:ttsvd}
        
        \Input{
            Data tensor, $\mathcal{X} \in \mathbb{R}^{I_1 \times I_2 \times \cdots \times I_N}$, and approximation accuracy, $\epsilon$
        }
    
        \Output{
            Core tensors, $\mathcal{G}^{(1)}, \dots, \mathcal{G}^{(N)}$, approximating $\mathcal{X} \in
            \mathbb{R}^{I_1 \times I_2 \times \cdots \times I_N}$
        }
    
        Initialize cores, $\mathcal{G}^{(1)}, \dots, \mathcal{G}^{(N)}$, and $R_0=1$
    
        Compute truncation parameter $\delta = \frac{\epsilon}{\sqrt{N-1}} ||\mathcal{X}||_F$
    
        $\mathcal{Z} \leftarrow \mathcal{X}$, and $\mathbf{Z} \leftarrow \mathbf{Z}_{(1)}$
    
        \For{$n=1$ to $N-1$}{
            Compute $\delta$-truncated SVD:
            $
                \mathbf{Z} = \mathbf{USV} + \mathbf{E},
                \text{ s.t. }
                {|| \mathbf{E} ||_F} \leq \delta; \mathbf{U} \in \mathbb{R}^{R_{(n-1)}I_n \times R_n}
            $
    
            $\mathcal{G}^{(n)} \leftarrow$
                \texttt{reshape} {$\left( \mathbf{U}, [ R_{(n-1)}, I_n, R_n   ] \right)$}
    
            $\mathbf{Z} \leftarrow$
                \texttt{reshape} {$\left( \mathbf{SV}^T, [R_n I_{(n+1)}, I_{(n+2)} I_{(n+3)}\dots I_N]) \right)$}
        }
    
        $\mathcal{G}^{(N)} \leftarrow \mathbf{Z}$
    
        \Return{
             $\mathcal{G}^{(1)}, \mathcal{G}^{(2)}, \dots, \mathcal{G}^{(N)}$
        }    
    \end{algorithm}

\subsection{Tensors and Tensor Operations}
This section gives brief mathematical preliminaries of tensor algebra, and basic knowledge in LLMs to facilitate the understanding of our proposed methodology in \cref{sec:methodology}.
\paragraph{Order-N Tensor.} 

    An order-$N$ real-valued tensor is a multi-dimensional array, denoted by a calligraphic font, e.g., $\mathcal{A}\in\mathbb{R}^{I_1\times\dots\times I_N}$, where $N$ is the order of the tensor (i.e., number of modes), and $I_n$ ($1 \leq n \leq N$) is the size (i.e., the dimension) of its $n$-{th} mode. Matrices (denoted by bold capital letters, e.g., $\mathbf{A}\in\mathbb{R}^{I_1\times I_2}$) can be seen as order-$2$ tensors ($N=2$), vectors (denoted by bold lower-case letters, e.g., $\mathbf{a}\in\mathbb{R}^{I}$) can be seen as order-1 tensors ($N=1$), and scalars (denoted by lower-case letters, e.g., $a\in\mathbb{R}$) are order-$0$ tensors ($N=0$).
    
\paragraph{Tensor Entries.} 
    
    The $(i_1, \ldots, i_N)$-th entry of an order-$N$ tensor is denoted by $a_{i_1, \cdots, i_N} \in \mathbb{R}$, where $i_n = 1, \ldots, I_n$ for $n=1,\ldots,N$. A tensor fiber is a vector of tensor entries obtained by fixing all but one index of the original tensor (e.g., $\mathbf{a}_{:, i_2, i_3, \ldots, i_N} \in \mathbb{R}^{I_1}$). Similarly, a tensor slice is a matrix of tensor entries obtained by fixing all but two indices of the original tensor (e.g., $\mathbf{A}_{:, :, i_3, i_4, \ldots, i_N} \in \mathbb{R}^{I_1 \times I_2}$). 

\paragraph{Tensorization.} 

    A vector, $\mathbf{a} \in \mathbb{R}^{I_1 I_2 \cdots I_N}$, can be tensorized (or {folded}) into an order-$N$ tensor, $\mathcal{A} \in \mathbb{R}^{I_1 \times I_2 \times \cdots \times I_N}$, with the relation between their entries defined by
    \begin{equation}\label{ttcp_eq:folding}
        \mathcal{A}[i_1, i_2, \dots, i_N] = a_i
    \end{equation}
    for all $1\leq i_n \leq I_n$, where $i=1+\sum_{n=1}^{N}(i_n-1)\prod_{k=1}^{n-1}I_k$.

\paragraph{Matricization (Mode-n unfolding).} 

    Mode-$n$ matricization of a tensor, $\texttt{mat}\left( \mathcal{A}, n \right) = \mathbf{A}_{\{n\}} \in \mathbb{R}^{I_n \times (I_1 \cdots I_{n-1} I_{n+1} \cdots I_N)}$, is a procedure of mapping the elements from a multidimensional array to a two-dimensional array (matrix). Conventionally, such procedure is associated with stacking mode-$n$ fibers (modal vectors) as column vectors of the resulting matrix. For instance, the mode-$1$ unfolding of $\mathcal{A} \in \mathbb{R}^{I_1 \times I_2 \times \cdots \times I_N}$ is represented as $\texttt{mat}\left( \mathcal{A}, 1 \right) = \mathbf{A}_{\{1\}} \in \mathbb{R}^{I_1 \times (I_2 \cdots I_N)}$, where the subscript, $\{1\}$, denotes the mode of matricization, and is given by
    \begin{equation}
        \mathbf{A}_{(1)}\bigg[i_1,\overline{i_2 i_3 \ldots i_N} \bigg] = \mathcal{A}[i_1,i_2,\ldots, i_N]
    \end{equation}
    Note that the overlined subscripts refer to linear indexing (or
    Little-Endian), given by:
    
    \begin{equation}\label{ttcp_eq:mode-n-unfold}
    \begin{aligned}
        \overline{i_1 i_2 \dots i_N}
            &= 1 + \sum_{n=1}^N \Bigg[ (i_n - 1) \prod_{n'=1}^{n-1}I_{n'} \Bigg] \\
            &= 1 + i_1 + (i_2 - 1)I_1 + \cdots + (i_n-1)I_1 \ldots I_{N-1}
    \end{aligned}
    \end{equation}

\paragraph{Tensor contraction.}

    The \textit{contraction} of an order-$N$ tensor, $\mathcal{A}\in\mathbb{R}^{I_1\times \dots \times I_N}$, and an order-$M$ tensor $\mathcal{B}\in\mathbb{R}^{J_1\times \dots \times J_M}$, over the $n$\textsuperscript{th} and $m$\textsuperscript{th} modes respectively, where $I_n=J_m$, results in $\mathcal{C} \in \mathbb{R}^{I_1 \times \cdots \times I_{n-1} \times I_{n+1} \times \cdots \times I_N \times J_1 \times \cdots \times J_{m-1} \times J_{m+1} \times J_M}$, with entries 
    \begin{equation}\label{ttcp_eq:cont}
    \begin{aligned}
        &c_{i_1,\dots,i_{n-1}, i_{n+1}, \dots, i_N, j_1, \dots, j_{m-1}, j_{m+1}, \dots, j_M   } = \sum_{i_n=1}^{I_n} a_{i_1, \dots, i_{n-1}, i_n, i_{n+1}, \dots, i_N} b_{j_1, \dots, j_{m-1}, i_n, j_{m+1}, \dots, j_M}
    \end{aligned}
    \end{equation}
    A $(2, 1)$-tensor contraction between two order-$2$ tensors, $\textbf{A} \in \mathbb{R}^{I_1 \times I_2}$ and $\textbf{B} \in \mathbb{R}^{J_1 \times J_2}$, where $I_2 = J_1$, is equivalent to a standard matrix multiplication, $\textbf{C} = \textbf{A} \times_2^1 \textbf{B} = \textbf{A} \textbf{B} \in \mathbb{R}^{I_1 \times J_2}$. Similarly, a $(2, 1)$-tensor contraction between an order-$2$ tensor, $\textbf{A} \in \mathbb{R}^{I_1 \times I_2}$, and an order-$1$ tensor, $\textbf{b} \in \mathbb{R}^{J_1}$, where $I_2 = J_1$, is equivalent to a standard matrix-by-vector multiplication, $\textbf{c} = \textbf{A} \times_2^1 \textbf{b} = \textbf{A} \textbf{b} \in \mathbb{R}^{I_1}$.

\paragraph{Tensor-Train Singular Value Decomposition (TT-SVD)}\label{app:ttsvd} is clarified in \cref{alg:ttsvd}.

\section{Proof of the Highest Compression Ratio in~\cref{tab:complexity}}\label{app:2-power}

Setting the tensor size $\left[I_1,\ldots,I_{N}\right]$ for the tensor $\mathcal{X}$ to achieve the highest compression rate, we next give the proof of this hyperparameter selection.

Regarding the definition of the compression rate in~\cref{sec:exp}, and $r_0=r_N=1$ in~\cref{sec:med-1}, the compression rate can be represented as
\begin{align}
 \eta &= \frac{V\times d} 
 {\sum_{j=1}^{V}\sum_{n=1}^{N}(r_{n-1} \times I_n \times r_n)_j} \\
 &= \frac{V \times d}{I_{1}r_{1}+r_{1}I_{2}r_{2}+\cdots+r_{N-2}I_{N-1}r_{N-1}+r_{N-1}I_{N}} \\
 & = \frac{V\times d}{\sum_{k=1}^{\lfloor \frac{N+1}{2} \rfloor} r_{2k-1} \left( r_{2k-2}I_{2k-1}+I_{2k}r_{2k+1} \right)}
\end{align}

For the simplest case, assume $I_1=\cdots=I_{N}=I$ and $r_1=\cdots=r_N=r$. Given $d=\prod_{n=1}^{N} I_n=I^{N}$, we have $N=\log_{I}{D}$, and
\begin{align}
    \eta &= \frac{V\times d}{rI\left[2+(N-2)r\right]}\\
    &=\frac{V\times d}{rI\left[2+(\log_{I} d-2)\right]}.\label{eq:com-app}
\end{align}

In Equation~\ref{eq:com-app}, the numerator is a constant, and in the denominator, $R$ is a hyperparameter for the Tensor-Train Decomposition. 
Thus the objective function for the highest compression rate $\eta$ is
\begin{align}
    \min_{I, N} rI\left[2+(N-2)\right] \quad \quad \textbf{s.t.} \quad  
    &N=\log_{I}{d} \label{eq:com-obj} \\ 
    & I, N, r \in \mathbb{Z}^{+} \\
    & 2\leq I \leq N \leq \lfloor\log_2 d\rfloor  \label{eq:com-range}
\end{align}

Regarding \cref{eq:com-obj}, if eliminate $N$ then we have a function $h=rI\left[2+(\log_{I}{d}-2)\right]$. Regarding $d$ in \cref{eq:com-range}, the largest token embedding size of recent GPT-3~\citep{brown2020language} is 12,888. Thus, for the GPT series models no later than GPT-3, \cref{eq:com-obj} should be  $2\leq I \leq N \leq 13$. In this range, $h$ is a monotonically increasing function, where the minimum $h$ occurs at $I=2$. 

Therefore, for the simplest case, we have the best hyperparameter selection of $I_1=I_2=\cdots=I_N=2$, $N=\lfloor\log_2 d\rfloor$.


\end{document}